# Image Compressive Sensing Recovery Using Adaptively Learned Sparsifying Basis via L0 Minimization

Jian Zhang[1*], Chen Zhao[2], Debin Zhao[1], Wen Gao[2]

[1]School of Computer Science and Technology, Harbin Institute of Technology, Harbin 150001, China
[2]School of Electrical Engineering and Computer Science, Peking University, Beijing 100871, China
e-mail: {jzhangcs, dbzhao}@hit.edu.cn; {zhaochen, wgao}@pku.edu.cn

*Abstract*—From many fewer acquired measurements than suggested by the Nyquist sampling theory, compressive sensing (CS) theory demonstrates that, a signal can be reconstructed with high probability when it exhibits sparsity in some domain. Most of the conventional CS recovery approaches, however, exploited a set of fixed bases (e.g. DCT, wavelet and gradient domain) for the entirety of a signal, which are irrespective of the non-stationarity of natural signals and cannot achieve high enough degree of sparsity, thus resulting in poor CS recovery performance. In this paper, we propose a new framework for image compressive sensing recovery using adaptively learned sparsifying basis via L0 minimization. The intrinsic sparsity of natural images is enforced substantially by sparsely representing overlapped image patches using the adaptively learned sparsifying basis in the form of L0 norm, greatly reducing blocking artifacts and confining the CS solution space. To make our proposed scheme tractable and robust, a split Bregman iteration based technique is developed to solve the non-convex L0 minimization problem efficiently. Experimental results on a wide range of natural images for CS recovery have shown that our proposed algorithm achieves significant performance improvements over many current state-of-the-art schemes and exhibits good convergence property.

## 1. Introduction

As a fundamental problem in the field of image processing, image restoration has been extensively studied in the past two decades [1]–[18]. It aims to reconstruct the original high quality image from its degraded observed version. It has been widely recognized that image prior knowledge plays a critical role in the performance of image restoration algorithms. Therefore, designing effective regularization terms to reflect the image priors is at the core of image restoration.

Classical regularization terms utilize local structural patterns and are built on the assumption that images are locally smooth except at edges. Several representative works in the literature are half quadrature formulation [2], Mumford-Shah (MS) model [3], and total variation (TV) models [1]. In recent years, very impressive image processing and restoration results have been obtained with local patch-based sparse representations calculated with dictionaries learned from natural images [11]–[14]. The sparse model assumes that each patch of an image can be accurately represented by a few elements from a basis set called a dictionary, which is learned from natural images [12]. Compared with traditional analytically-designed dictionaries, such as wavelets, curvelets, and bandlets, the learned dictionary enjoys the advantage of being better adapted to the images, thereby enhancing the sparsity and showing impressive performance improvement [13], [14]. Another alternative significant property exhibited in natural images is the well-known nonlocal self-similarity, which depicts the repetitiveness of higher level patterns (e.g., textures and structures) globally positioned in images. A representative work is the popular nonlocal means (NLM) [5], which takes advantage of this image property to conduct a type of weighted filtering for denoising tasks by means of the degree of similarity among surrounding pixels. Later, inspired by the success of nonlocal means (NLM) denoising filter, a series of nonlocal regularization terms for inverse problems exploiting nonlocal self-similarity property of natural images are emerging [10], [15]. In recent works, the sparsity and the self-similarity of natural images are usually combined to achieve better performance [16]–[18].

In the field of image restoration, perhaps the hottest topic is the recent development of Compressive Sensing (CS) theory, which has drawn quite an amount of attention as an alternative to the current methodology of sampling followed by compression [19]–[21]. By exploiting the redundancy existed in a signal, CS conducts sampling and compression at the same time. CS theory shows that a signal can be decoded from many fewer measurements than suggested by the Nyquist sampling theory, when the signal is sparse in some domain, which has greatly changed the way engineers think of data acquisition.

In CS theory, a signal is usually sampled by a random projection that is signal-independent and reconstructed by minimizing the $\ell_0$ or $\ell_1$ optimization problem with the prior that the signal is sparse in some transformation domain. Since the $\ell_0$ minimization is discontinuous and an NP-hard problem, the most common one is to use the $\ell_1$ norm, which is the optimal convex approximation of $\ell_0$ norm and has been proved that for many problems it is probable that the $\ell_1$ norm is equivalent to the $\ell_0$ norm in a technical sense. This equivalence result allows one to solve the $\ell_1$ problem, which is easier than the $\ell_0$ problem. Many CS recovery algorithms are recently proposed: linear programming [22], gradient projection sparse reconstruction [23], matching pursuit [24], and iterative thresholding [25].

An attractive strength of CS is that the sampling process is made signal-independent and computationally inexpensive at the cost of high reconstruction complexity. This asymmetric design is severely desirable in some image processing applications when the data acquisition devices must be simple (e.g. inexpensive resource-deprived sensors) [26], or when oversampling can harm the object being captured (e.g. X-ray imaging) [27].

CS theory shows that the sparsity degree of a signal plays a significant role in recovery. The higher degree of a signal, the higher recovery quality it will have. So, seeking a domain in which the signal has a high degree of sparsity is one of the main challenges CS recovery should face. However, natural signals such as images are typically non-stationary, there exists no universal domain in which all parts of the signals are sparse. The most current CS recovery methods explore a set of fixed domains (e.g. DCT, wavelet and gradient domain) [28, 29, 30, 34], and therefore are signal-independent or not adaptive, resulting in poor CS recovery performance.

Towards this problem, Wu et al. [27] proposed a model-guided adaptive recovery of compressive sensing (MARX) utilizing a piecewise autoregressive model to adapt to the changing second order statistics of natural images. Many recent works incorporated additional prior knowledge about transform coefficients (statistical dependencies, structure, etc.) into the CS recovery framework, such as Gaussian scale mixtures (GSM) models [31], tree-structured wavelet (TSW) [32], tree-structured DCT (TSDCT) [33].

Considering the fact that the natural image signal is non-stationary and inspired by the recent great success of sparse representation in image processing, in this paper, we propose to enforce the intrinsic sparsity of a natural image by sparsely representing its overlapped image patches using adaptively learned sparsifying basis. All the sparse codes of image patches constitute the adaptive redundant sparse representation of the whole image, which is incorporated into the optimization problem for the whole image CS recovery in the form of $\ell_0$ norm, greatly reducing blocking artifacts and confining the CS solution space. In addition, to make our proposed scheme tractable and robust, a split Bregman iteration based technique is developed to solve the non-convex $\ell_0$ minimization problem efficiently. Experimental results on a wide range of natural images for CS recovery have shown that our proposed algorithm is quite competitive to the state-of-the-art methods.

The remainder of the paper is organized as follows. Section 2 briefly reviews CS theory and introduce split Bregman iteration algorithm. Section 3 provides our proposed framework for image CS recovery using adaptive learned sparsifying basis via $\ell_0$ minimization. The implementation details of optimization are elaborated in Section 4. Experimental results are reported in Section 5. In Section 6, we conclude this paper.

## 2. Background

*2.1 Compressive Sensing*

When the signal is sparse in some domain, CS allows exact recovery of the signal from its measurements acquired by linear projection, whose number is much smaller than that of the original signal. Suppose a signal $\boldsymbol{x} \in \mathbb{R}^N$ is of size *N*, and its transform coefficient vector over domain $\Psi$ is denoted by $\theta$, i.e. $\boldsymbol{x} = \Psi\theta$. $\boldsymbol{x}$ is said to be sparse in $\Psi$, if the elements in $\theta$, are mostly zeros, or nearly sparse if the dominant portion of $\theta$ are either zeros or very close to zeros. The sparsity of $\boldsymbol{x}$ in $\Psi$ is quantified by the number of significant elements within the coefficient vector $\theta$.

More specifically, denote the linear measurements of $\boldsymbol{x}$ by $\boldsymbol{y} \in \mathbb{R}^M$, namely, $\boldsymbol{y} = \Phi\boldsymbol{x}$. Here, $\Phi$ represents the random projections and is an *M*×*N* measurement matrix such that *M* is much smaller than *N*. The purpose of CS recovery is to recover $\boldsymbol{x}$ from $\boldsymbol{y}$ with subrate, being *S = M/N*, which is usually formulated as the following $\ell_p$ optimization problem:

$$\min_{\theta} \tfrac{1}{2} \| \boldsymbol{y} - \Phi\Psi\theta \|_2^2 + \lambda \| \theta \|_p \quad , \tag{1}$$

where $\lambda$ is non-negative parameter, and $p$ is usually set to 1 or 0, characterizing the sparsity of the vector $\theta$. $\|*\|_1$ is $\ell_1$ norm, adding all the absolute values of the entries in a vector, while $\|*\|_0$ is $\ell_0$ norm, counting the nonzero entries of a vector. According to [19], CS is capable of recovering *K*-sparse signal $\boldsymbol{x}$ (with an overwhelming probability) from $\boldsymbol{y}$ of size *M*, provided that the number of random samples meets *M ≥ cK(N/K)*. The required sampling rate (*M/K*), to incur lossless recovery, is roughly proportional to (*K/N*). A compressive imaging camera prototype using random projection has been presented in [21].

*2.2 Split Bregman Iteration (SBI)*

In order to facilitate the discussions in the following optimization section, this section briefly introduces the well-known convex optimization algorithm split Bregman iteration (SBI). Split Bregman Iteration (SBI) is recently introduced by [36] for solving a class of $\ell_1$ related minimization problems. The basic idea of SBI is to convert the unconstrained minimization problem into a constrained one by introducing the variable splitting technique and then invoke the Bregman iteration [35] to solve the constrained minimization problem. Numerical simulations in show that it converges fast and only uses a small memory footprint, which makes it very attractive for large-scale problems [37].

Consider a constrained optimization problem

$$\min_{\boldsymbol{u}\in\mathbb{R}^N, \boldsymbol{v}\in\mathbb{R}^M} f(\boldsymbol{u}) + g(\boldsymbol{v}), \quad \text{s.t.} \quad \boldsymbol{u} = \boldsymbol{G}\boldsymbol{v}, \tag{2}$$

where $\boldsymbol{G} \in \mathbb{R}^{N \times M}$ and $f: \mathbb{R}^N \to \mathbb{R}$, $g: \mathbb{R}^M \to \mathbb{R}$ are convex functions. The SBI to address problem (2) works as follows:

**Algorithm 1** *Split Bregman Iteration (SBI)*

1. **Set** $t = 0$, choose $\mu > 0$, $\boldsymbol{b}_0 = \boldsymbol{0}, \boldsymbol{u}_0 = \boldsymbol{0}, \boldsymbol{v}_0 = \boldsymbol{0}$.
2. **Repeat**
3. $\boldsymbol{u}^{(t+1)} = \operatorname{argmin}_u f(\boldsymbol{u}) + \frac{\mu}{2} \|\boldsymbol{u} - \boldsymbol{G}\boldsymbol{v}^{(t)} - \boldsymbol{b}^{(t)}\|_2^2$;
4. $\boldsymbol{v}^{(t+1)} = \operatorname{argmin}_v g(\boldsymbol{v}) + \frac{\mu}{2} \|\boldsymbol{u}^{(t+1)} - \boldsymbol{G}\boldsymbol{v} - \boldsymbol{b}^{(t)}\|_2^2$;
5. $\boldsymbol{b}^{(t+1)} = \boldsymbol{b}^{(t)} - (\boldsymbol{u}^{(t+1)} - \boldsymbol{G}\boldsymbol{v}^{(t+1)})$;
6. $t \leftarrow t + 1$;
7. **Until** stopping criterion is satisfied

In SBI, the parameter $\mu$ is fixed to avoid the problem of numerical instabilities instead of choosing a predefined sequence $\{\mu_k\}$ that tends to infinity. The convergence of Split Bregman Iteration can be attested by the equivalence between SBI and the Douglas-Rachford splitting method (DRSM) applied to its dual problem [38].

## 3. Image CS recovery using adaptively learned sparsifying basis via L0 minimization

In this section, we first introduce the patch-based redundant sparse representation of natural images, and then establish a new framework for image compressive sensing recovery using adaptively learned sparsifying basis via $\ell_0$ minimization.

*3.1 Patch-based Redundant Sparse Representation*

In literature, the basic unit of sparse representation for natural images is patch [12]. Mathematically, denote by $\boldsymbol{x} \in \mathbb{R}^N$ and $\boldsymbol{x}_k \in \mathbb{R}^{B_s}$ the vector representations of the original image and an image patch of size $\sqrt{B_s} \times \sqrt{B_s}$ at location $k$, $k = 1, 2, ..., n$. Then we have

$$\boldsymbol{x}_k = \boldsymbol{R}_k \boldsymbol{x}, \tag{3}$$

where $\boldsymbol{R}_k \in \mathbb{R}^{B_s \times N}$ is a matrix operator that extracts patch $\boldsymbol{x}_k$ from $\boldsymbol{x}$. Note that patches are usually overlapped, and such patch-based representation is highly redundant. In the experimental section, we will illustrate that the overlapped technique and the patch-based redundant representation are significant to achieve high recovery quality. Therefore, the recovery of $x$ from $\{x_k\}$ becomes an over-determined system, which is straightforward to obtain the following Least-Square solution [11]:

$$x = \left(\sum_{k=1}^{n} R_k^T R_k\right)^{-1} \sum_{k=1}^{n} \left(R_k^T x_k\right), \tag{4}$$

which is nothing but an abstraction strategy of averaging all the overlapped patches.

Given dictionary $D \in \mathbb{R}^{B_s \times M}$, the sparse coding process of each patch $x_k$ over $D$ is to find a sparse vector $\alpha_k$ (i.e., most of the coefficients in $\alpha_k$ are zero or close to zero) such that $x_k \approx D\alpha_k$. Then the entire image can be sparsely represented by the set of sparse codes $\alpha_k$.

Similar to Eq. (4), reconstructing $x$ from its sparse codes $\alpha_k$ is formulated:

$$x \approx D \circ \alpha = \left(\sum_{k=1}^{n} R_k^T R_k\right)^{-1} \sum_{k=1}^{n} \left(R_k^T D\alpha_k\right), \tag{5}$$

where $\alpha$ denotes the concatenation of all $\alpha_k$, that is, $\alpha = [\alpha_1^T, \alpha_2^T, ..., \alpha_n^T]^T$, which is patch-based redundant sparse representation for $x$.

*3.2 Image CS Recovery via $\ell_0$ Minimization*

Now, incorporating Eq. (5) into Eq. (1), our proposed scheme for image compressive sensing recovery using adaptive learned sparsifying basis via $\ell_0$ minimization is formulated as:

$$\min_{\alpha} \tfrac{1}{2} \| y - \Phi D \circ \alpha \|_2^2 + \lambda \| \alpha \|_0 \ . \tag{6}$$

Here, $D$ replaces $\Psi$ in Eq. (1), standing for adaptively learned sparsifying basis, which will be given in the next section, and $\alpha$ denotes the patch-based redundant sparse representation for the whole image over $D$.

As we know, since $\ell_0$ minimization is non-convex and NP-hard, the usual routine is to solve its optimal convex approximation, i.e., $\ell_1$ minimization, which has been proved that, under some conditions, $\ell_1$ minimization is equivalent to $\ell_0$ minimization in a technical sense. The $\ell_1$ minimization can be solved efficiently by some recent convex optimization algorithms, such as iterative shrinkage/thresholding [39], [40], [45], and split Bregman [36], [37] algorithms. Therefore, the straightforward method to solve Eq. (6) is translated into solving its $\ell_1$ convex form, that is

$$\min_{\alpha} \tfrac{1}{2} \| y - \Phi D \circ \alpha \|_2^2 + \lambda \| \alpha \|_1 \ . \tag{7}$$

However, a fact that is often neglected is, for some practical problems including image inverse problems, the conditions guaranteeing the equivalence of $\ell_0$ minimization and $\ell_1$ minimization are not necessarily satisfied. Consequently, this paper proposes to exploit the framework of convex optimization algorithms to solve the non-convex $\ell_0$ minimization, i.e., Eq. (6) directly. Experimental results demonstrate the effectiveness and the convergence of our proposed approach.

**4. Optimization for Proposed L0 minimization**

In this paper, we adopt the framework of split Bregman iteration (SBI) [37] to solve Eq. (6), which is verified to be more ef-fective than iterative shrinkage/thresholding (IST) [39] in our experiments (See Section 5 for more details). The developed optimization details to solve the proposed $\ell_0$ minimization problem effectively and efficiently are given in this section.

According to SBI in Section 2, the original minimization problem (2) is split into two sub-problems. The rationale behind is that each sub-problem minimization may be much easier than the original problem (2).

Now, let us go back to Eq. (6) and point out how to apply the framework of SBI to solve it. By introducing a variable $u$, we first transform Eq. (6) into an equivalent constrained form, i.e.,

$$\min_{\alpha,u} \tfrac{1}{2}\|y-\Phi u\|_2^2 + \lambda\|\alpha\|_0, \text{ s.t. } u = D \circ \alpha. \tag{8}$$

Define $f(u) = \tfrac{1}{2}\|y-\Phi u\|_2^2$, $g(\alpha) = \lambda\|\alpha\|_0$.

Then, invoking SBI, Line 3 in **Algorithm 1** becomes:

$$u^{(t+1)} = \arg\min_u \tfrac{1}{2}\|y-\Phi u\|_2^2 + \tfrac{\mu}{2}\|u - D \circ \alpha^{(t)} - b^{(t)}\|_2^2. \tag{9}$$

Next, Line 4 in **Algorithm 1** becomes:

$$\alpha^{(t+1)} = \arg\min_\alpha \lambda\|\alpha\|_0 + \tfrac{\mu}{2}\|u^{(t+1)} - D \circ \alpha - b^{(t)}\|_2^2. \tag{10}$$

According to Line 5 in **Algorithm 1**, the update of $b^{(t)}$ is

$$b^{(t+1)} = b^{(t)} - (u^{(t+1)} - D \circ \alpha^{(t+1)}). \tag{11}$$

Thus, by SBI, the minimization for Eq. (6) is transformed into solving two sub-problems, namely, $u$, $\alpha$ sub-problems. In the following, we will provide the implementation details to obtain the efficient solutions to each separated sub-problem. For simplicity, the subscript $t$ is omitted without confusion.

*4.1 $u$ Sub-problem*

Given $x$, the $u$ sub-problem denoted by Eq. (9) is essentially a minimization problem of strictly convex quadratic function, that is

$$\min_u Q_1(u) = \min_u \tfrac{1}{2}\|y-\Phi u\|_2^2 + \tfrac{\mu}{2}\|u - D \circ \alpha - b\|_2^2. \tag{12}$$

Setting the gradient of $Q_1(u)$ to be zero gives a closed solution for Eq. (12), which can be expressed as

$$\hat{u} = (\Phi^T\Phi + \mu I)^{-1} q, \tag{13}$$

where $q = \Phi^T y + \mu(D \circ \alpha + b)$, $I$ is identity matrix.

As for image compressive sensing (CS) recovery, however, $\Phi$ is a random projection matrix without special structure. Thus, it is too costly to solve Eq. (12) directly by Eq. (13). Here, to avoid computing the matrix inverse, the gradient descent method is utilized to solve Eq. (12) by applying

$$\hat{u} = u - \eta d, \tag{14}$$

where $d$ is the gradient direction of the objective function $Q_1(u)$ and $\eta$ represents the step. Therefore, solving $u$ sub-problem for image CS recovery only requires computing the following equation iteratively

$$\hat{u} = u - \eta\left(\Phi^T\Phi u - \Phi^T y + \mu(u - D\circ\alpha - b)\right). \tag{15}$$

where $\Phi^T\Phi$ and $\Phi^T y$ can be calculated before, making above computation more efficient. As a matter of fact, in our implementation, one iteration is accurate enough.

*4.2 $\alpha$ Sub-problem*

Given $u$, according to Eq. (10), the $\alpha$ sub-problem can be formulated as

$$\min_\alpha Q_2(\alpha) = \min_\alpha \tfrac{1}{2}\|D\circ\alpha - r\|_2^2 + \tfrac{\lambda}{\mu}\|\alpha\|_0, \tag{16}$$

where $r = u - b$.

Note that it is difficult to solve Eq. (16) directly due to the complicated definition of $\alpha$. Instead, we make some transformation. Let $x = D\circ\alpha$, then Eq. (16) equally becomes

$$\min_\alpha \tfrac{1}{2}\|x - r\|_2^2 + \tfrac{\lambda}{\mu}\|\alpha\|_0. \tag{17}$$

To enable a tractable solution to Eq. (17), in this paper, a general assumption is made. Concretely, we regard $r$ as some type of the noisy observation of $x$, denote the error vector by $e = x - r$, and then make an assumption that each element of $e$ follows an independent zero-mean distribution with the same variance $\sigma^2$. It is worth emphasizing that the above assumption does not need to be Gaussian process, which is more general and reasonable. By this assumption, we can prove the following conclusion.

**THEOREM 1.** *Let $x, r \in \mathbb{R}^N, x_k, r_k \in \mathbb{R}^{B_s}$, and denote the error vector by $e = x - r$ and each element of $e$ by $e(j)$, $j = 1,...,N$. Assume that $e(j)$ is independent and comes from a distribution with zero mean and variance $\sigma^2$. Then, for any $\varepsilon > 0$, we have the following property to describe the relationship between $\|x - r\|_2^2$ and $\sum_{k=1}^n \|x_k - r_k\|_2^2$, that is,*

$$\lim_{\substack{N\to\infty \\ K\to\infty}} P\left\{\left|\tfrac{1}{N}\|x - r\|_2^2 - \tfrac{1}{K}\sum_{k=1}^n \|x_k - r_k\|_2^2\right| < \varepsilon\right\} = 1, \tag{18}$$

*where $P(\cdot)$ represents the probability and $K = B_s \times n$.*
*Proof:*

Due to the assumption that each $e(j)$ is independent, we obtain that each $e(j)^2$ is also independent. Since $E[e(j)] = 0$ and $\text{Var}[e(j)] = \sigma^2$, we have the mean of each $e(j)^2$, which is expressed as

$$E[e(j)^2] = \text{Var}[e(j)] + [E[e(j)]]^2 = \sigma^2, \quad j = 1,...,N. \tag{19}$$

By invoking *Law of Large Numbers* in probability theory, for any $\varepsilon > 0$, it yields $\lim_{N\to\infty} P\left\{\left|\tfrac{1}{N}\sum_{j=1}^N e(j)^2 - \sigma^2\right| < \tfrac{\varepsilon}{2}\right\} = 1$, i.e.,

$$\lim_{N\to\infty} P\left\{\left|\tfrac{1}{N}\|x - r\|_2^2 - \sigma^2\right| < \tfrac{\varepsilon}{2}\right\} = 1, \tag{20}$$

Further, let $\boldsymbol{x}_P, \boldsymbol{r}_P$ denote the concatenation of all the patches $\boldsymbol{x}_k$ and $\boldsymbol{r}_k$, $k=1,2,...,n$, respectively, and denote each element of $\boldsymbol{x}_P - \boldsymbol{r}_P$ by $\boldsymbol{e}_P(i), i=1,...,K$. Due to the assumption, we conclude that $\boldsymbol{e}_P(i)$ is independent with zero mean and variance $\sigma^2$.

Therefore, the same manipulations with Eq. (20) applied to $\boldsymbol{e}_P(i)^2$ lead to $\lim_{K\to\infty} P\left\{\left|\frac{1}{K}\sum_{i=1}^{K}\boldsymbol{e}_P(i)^2 - \sigma^2\right| < \frac{\varepsilon}{2}\right\} = 1$, namely,

$$\lim_{K\to\infty} P\left\{\left|\frac{1}{K}\sum_{k=1}^{n}\|\boldsymbol{x}_k - \boldsymbol{r}_k\|_2^2 - \sigma^2\right| < \frac{\varepsilon}{2}\right\} = 1 . \tag{21}$$

Considering Eqs. (20) and (21) together, we prove Eq. (18).

According to **Theorem 1**, there exists the following equation with very large probability (limited to 1) at each iteration $t$:

$$\frac{1}{N}\|\boldsymbol{x}^{(t)} - \boldsymbol{r}^{(t)}\|_2^2 = \frac{1}{K}\sum_{k=1}^{n}\|\boldsymbol{x}_k^{(t)} - \boldsymbol{r}_k^{(t)}\|_2^2 . \tag{22}$$

Next, by incorporating Eq. (22) into Eq. (17), it yields

$$\begin{aligned}
&\min_{\boldsymbol{\alpha}} \tfrac{1}{2}\sum_{k=1}^{n}\|\boldsymbol{x}_k - \boldsymbol{r}_k\|_2^2 + \tfrac{\lambda K}{\mu N}\|\boldsymbol{\alpha}\|_0 \\
&= \min_{\boldsymbol{\alpha}} \tfrac{1}{2}\sum_{k=1}^{n}\|\boldsymbol{x}_k - \boldsymbol{r}_k\|_2^2 + \tfrac{\lambda K}{\mu N}\sum_{k=1}^{n}\|\boldsymbol{\alpha}_k\|_0 \\
&= \min_{\boldsymbol{\alpha}} \sum_{k=1}^{n} \tfrac{1}{2}\|\boldsymbol{x}_k - \boldsymbol{r}_k\|_2^2 + \tau\|\boldsymbol{\alpha}_k\|_0 ,
\end{aligned} \tag{23}$$

where $\tau = (\lambda K)/(\mu N)$.

It is obvious to see that Eq. (23) can be efficiently minimized by solving $n$ sub-problems for all the overlapped patches $\boldsymbol{x}_k$. Each patch based sub-problem is formulated as:

$$\begin{aligned}
&\operatorname{argmin}_{\boldsymbol{\alpha}_k} \tfrac{1}{2}\|\boldsymbol{x}_k - \boldsymbol{r}_k\|_2^2 + \tau\|\boldsymbol{\alpha}_k\|_0 \\
&= \operatorname{argmin}_{\boldsymbol{\alpha}_k} \tfrac{1}{2}\|\boldsymbol{D}\boldsymbol{\alpha}_k - \boldsymbol{r}_k\|_2^2 + \tau\|\boldsymbol{\alpha}_k\|_0
\end{aligned} . \tag{24}$$

Obviously, Eq. (24) can also be considered as the sparse coding problem. To achieve high sparsity, we directly solve the constrained form of Eq. (24), i.e.,

$$\min_{\boldsymbol{\alpha}_k} \|\boldsymbol{\alpha}_k\|_0 \quad \text{s.t.} \quad \|\boldsymbol{D}\boldsymbol{\alpha}_k - \boldsymbol{r}_k\|_2^2 \leq \delta , \tag{25}$$

where $\omega$ is a control factor and $\delta = \omega\tau$.

Note that Eq. (25) can be achieved efficiently by orthogonal matching pursuit (OMP) algorithm [24]. This process is applied for all $n$ overlapped patches to achieve $\hat{\boldsymbol{\alpha}}$, which is the final solution for $\boldsymbol{\alpha}$ sub-problem in Eq. (16).

*4.3 Adaptive Sparsifying Basis Learning*

The key of the sparse representation modeling lies in the choice of dictionary or sparsifying basis $\boldsymbol{D}$. In other words, how to seek the best domain to sparsify a given image? Much effort has been devoted to learning a redundant sparsifying basis from a set of training example image patches. To be concrete, given a set of training image patches $\boldsymbol{S} = [\boldsymbol{s}_1, \boldsymbol{s}_2, ..., \boldsymbol{s}_J]$, the goal of sparsifying basis learning is to jointly optimize the sparsifying basis $\boldsymbol{D}$ and the representation coefficients matrix $\boldsymbol{\Lambda} = [\boldsymbol{\alpha}_1, \boldsymbol{\alpha}_2, ..., \boldsymbol{\alpha}_J]$ such that $\boldsymbol{s}_k = \boldsymbol{D}\boldsymbol{\alpha}_k$ and $\|\boldsymbol{\alpha}_k\|_p \leq L$, where $p$ is 0 or 1. This can be formulated by the following minimization problem:

$$(\hat{\boldsymbol{D}}, \hat{\boldsymbol{\Lambda}}) = \underset{\boldsymbol{D}, \boldsymbol{\Lambda}}{\operatorname{argmin}} \sum\nolimits_{k=1}^{J} \left\| \boldsymbol{s}_k - \boldsymbol{D}\boldsymbol{\alpha}_k \right\|_2^2 \text{ s.t. } \left\| \boldsymbol{\alpha}_k \right\|_p \leq L, \forall k. \tag{26}$$

Apparently, the above minimization problem in Eq. (26) is large-scale and highly non-convex even when $p$ is 1. To make it tractable and solvable, some approximation approaches, including MOD [41] and K-SVD [12], have been proposed to optimize $\boldsymbol{D}$ and $\boldsymbol{\Lambda}$ alternatively, leading to many state-of-the-art results in image processing.

In order to achieve adaptive sparsifying basis, the training image patches usually come from the original image. Nonetheless, in practice, the original image $\boldsymbol{x}$ is not available, while we only have access to the CS measurements in Eq. (1). Such a problem with chicken-and-egg flavor is usually solved by an iterative way in which we obtain the estimate of $\boldsymbol{x}$ and $\boldsymbol{D}$ alternately [17]. Because $\boldsymbol{r}$ in Eq. (17) is regarded as a good approximation of $\boldsymbol{x}$ at each iteration, in this paper, we conduct adaptive sparsifying basis learning using all the patches extracted from $\boldsymbol{r}$. Due to its effectiveness and efficiency, K-SVD is adopted as the adaptive sparsifying basis learning method. More details about K-SVD can be found in [12].

*4.4 Summary of Proposed Algorithm*

So far, all issues in the process of handing the above two sub-problems have been solved. In fact, we acquire the efficient solution for each sbeparated sub-problem, which enables the whole algorithm more efficient and effective. In light of all derivations above, a detailed description of the proposed framework for image CS recovery using adaptive sparsifying basis via $\ell_0$ minimization is provided in Table 1.

## 5. Experimental results

In this section, experimental results are presented to evaluate the performance of our proposed framework for image CS recovery using adaptive sparsifying basis via $\ell_0$ minimization. Six test images are shown in Fig. 1. In our experiments, the CS measurements are obtained by applying a Gaussian random projection matrix to the original image signal at block level, i.e., block-based CS with block size of 32×32. The default parameter setting of proposed scheme is as follows: the size of each patch, i.e., $\sqrt{B_s} \times \sqrt{B_s}$ is 8×8, and the size of sparsifying basis is 256. $\mu$ is set to be 2.5e-3, $\eta$ is set to be 1, and $\omega$ is set to be 2. The value of $\lambda$ is related to the overlapped step size, which will be given in the following. All the experiments are performed in Matlab 7.12.0 on a Dell OPTIPLEX computer with Intel(R) Core(TM) 2 Duo CPU E8400 processor (3.00GHz), 3.25G memory, and Windows XP operating system.

*5.1 Effect of Overlapped Step Size*

The overlapped step size is defined as the distance between two adjacent patches to be processed. If the overlapped step size is the same as the patch size, all the patches are non-overlapped. If the overlapped step size is equal to one, it means the difference between two adjacent patches in the horizontal (or vertical) direction is only one column (or row). We first discuss the effect of overlapped step size to the CS recovery quality.

In Fig. 2, the performance of various overlapped step sizes for two test images in the cases of image CS recovery with *subrate=30%* are provided. Obviously, the results illustrate that smaller overlapped step size provides higher quality of processed images. This is mainly because more estimates are generated for the image with smaller overlapped step sizes, which further demonstrates the effectiveness of the patch-based redundancy representation for natural images. In addition, the overlapped strategy takes advantage of the correlations between blocks to depress the blocking artifacts. Therefore, in the following experiments, the overlapped step size is set be to 1. Furthermore, we have the relationship $K=64N$. Accordingly, in our test the parameter $\lambda$ is empirically set to be 1.4e-3.

*5.2 Effect of Sparsifying Basis Selection*

In this sub-section, we will show the effect of sparsifying basis selection to the image CS recovery. Three types of sparsifying basis selections are given. The first one is to choose the fixed over-complete DCT basis, as shown in Fig. 3(a). The second one is to choose the global sparsifying basis, which is learned from a large set of natural images, as shown in Fig. 3(b). The last one is to choose the suggested sparsifying basis in our paper, which is adaptively learned from the processed image at each iteration. Fig. 3(c) shows the adaptively learned sparsifying basis for image *House* at last iteration. Fig. 4 provides the CS recovery results for image *House* in the case of *subrate =20%* using three different sparsifying basis. From Fig. 4, it is clear to see that the recovered result by globally learned sparsifying basis is better than the one by fixed over-complete DCT basis. However, the adaptively learned sparsifying basis produces the best result, preserving sharper edges and finer details, which verifies the superiority of the proposed adaptively learned sparsifying basis for image CS recovery.

*5.3 Comparison between SBI and IST*

In previous works [30], [42], the $\ell_0$ minimization non-convex optimizations for image CS recovery are usually solved by iterative hard-thresholding algorithm [40], [45], which can be regarded as a special type of iterative shrinkage/thresholding (IST) algorithm [39]. Specifically, consider the following general optimization problem

$$\min_{\boldsymbol{u} \in \mathbb{R}^N} f(\boldsymbol{u}) + g(\boldsymbol{u}), \tag{27}$$

where $f(\boldsymbol{u})$ is a smooth convex function with gradient which is Lipschitz continuous, and $g(\boldsymbol{u})$ is a continuous convex function which is possibly non-smooth. The IST algorithm to solve problem (27) with constant step $\rho$ is formulated as:

$$\boldsymbol{r}^{(t+1)} = \boldsymbol{u}^{(t)} - \rho \nabla f(\boldsymbol{u}^{(t)}), \tag{28}$$

$$\boldsymbol{u}^{(t+1)} = \arg\min_{\boldsymbol{u}} \frac{1}{2} \|\boldsymbol{u} - \boldsymbol{r}^{(t+1)}\|_2^2 + \lambda g(\boldsymbol{u}). \tag{29}$$

Then, applying IST to solve our proposed non-convex $\ell_0$ minimization Eq. (6) with constraint $\boldsymbol{u} = \boldsymbol{D} \circ \boldsymbol{\alpha}$ can be expressed as the following iterations:

$$\boldsymbol{u}^{(t)} = \boldsymbol{D}^{(t)} \circ \boldsymbol{\alpha}^{(t)}, \tag{30}$$

$$\boldsymbol{r}^{(t+1)} = \boldsymbol{u}^{(t)} - \rho \Phi^T (\Phi \boldsymbol{u}^{(t)} - \boldsymbol{y}), \tag{31}$$

$$\boldsymbol{\alpha}^{(t+1)} = \text{argmin}_{\boldsymbol{\alpha}} \tfrac{1}{2} \left\| \boldsymbol{D} \circ \boldsymbol{\alpha} - \boldsymbol{r}^{(t+1)} \right\|_2^2 + \lambda \|\boldsymbol{\alpha}\|_0. \tag{32}$$

It is obvious to observe that Eq. (32) is equivalent to the above $\boldsymbol{\alpha}$ sub-problem, which can be solved efficiently. Hence, it is tractable to address Eq. (6) with IST algorithm.

Note that applying the convex optimization algorithm SBI to solve our proposed non-convex $\ell_0$ minimization is one of the main contributions of this paper. Here, we make a comparison between SBI and IST in our proposed image CS recovery framework using adaptively learned sparsifying basis. Take the cases of image CS recovery with *subrate=30%* for two gray images *Leaves* and *Vessels* as examples. Fig. 5 plots their progression curves of the PSNR (dB) results achieved by solving proposed $\ell_0$ minimization with SBI and IST. The result achieved by proposed $\ell_0$ minimization with SBI is denoted by SBI (red solid line), while the result achieved by proposed $\ell_0$ minimization with IST is denoted by IST (black dotted line). Obviously, SBI is more efficient and effective to solve our proposed $\ell_0$ minimization problem than IST, with more than 2 dB and 3 dB gains for images *Leaves* and *Vessels*. The visual comparison of image CS recovery between SBI and IST for image *Leaves* is shown in Fig. 6. It is apparent that the magnitude of the image CS recovery error (with respect to the original image) by SBI is much lower than that by IST, which fully demonstrates the superiority of SBI over IST for solving the proposed non-convex $\ell_0$ minimization.

*5.4 Comparison with State-of-the-Art Algorithms*

Our proposed algorithm is compared with four representative CS recovery methods in literature, i.e., wavelet method (DWT) [30], total variation (TV) method [29], multi-hypothesis (MH) method [42], collaborative sparsity (CoS) method [43], which deal with image signals in the wavelet domain, the gradient domain, the random projection residual domain, and the hybrid space-transform domain, respectively. It is worth emphasizing that MH and CoS are known as the current state-of-the-art algorithms for image CS recovery. **Due to the limit of space, only parts of the experimental results are shown in this paper. Please enlarge and view the figures on the screen for better comparison.** Our Matlab code and all the experimental results can be downloaded at the website: http://idm.pku.edu.cn/staff/zhangjian/ALSB/.

To evaluate the quality of the reconstructed image, in addition to PSNR (Peak Signal to Noise Ratio, unit: dB), which is used to evaluate the objective image quality, a recently proposed powerful perceptual quality metric FSIM (Feature SIMilarity) [44] is calculated to evaluate the visual quality. The higher FSIM value means the better visual quality. The PSNR and FSIM comparisons for six gray test images in the cases of *20% to 40%* measurements are provided in Table 2 and Table 3, respectively. Our proposed algorithm achieves the highest PSNR and FSIM among the five comparative algorithms in most cases, which can improve roughly 6.2 dB, 5.5 dB, 2.6 dB, and 2.0 dB on average, in comparison with DWT, TV, MH, CoS, respectively, greatly improving existing image CS recovery results.

Some visual results of the recovered images by various algorithms are presented in Figs. 7~11. Obviously, DWT and TV generate the worst perceptual results. The CS recovered images by MH and CoS possess much better visual quality than those of DWT and TV, but still suffer from some undesirable artifacts, such as ringing effects and lost details. The proposed algorithm not only eliminates most of the ringing effects, but also preserves much sharper edges and finer details, showing much clearer and better visual results than the other competing methods. The high performance is attributed to the proposed $\ell_0$ minimization, offering a powerful mechanism of characterizing the intrinsic sparsity of natural images by the redundant patch-based sparse representation using adaptively learned sparsifying basis, which is further solved efficiently by the proposed SBI based iterative techniques. Our work also offers a fresh and successful instance to corroborate the CS theory applied for natural images.

*5.5 Algorithm Complexity and Computational Time*

The complexity of the proposed algorithm is provided as follows. The solution to Problem (26) involves learning the adaptive sparsifying basis from a fraction of all *N* patches and exploiting it to obtain sparse approximations of the *N* patches. The sparsifying basis step utilizes the K-SVD algorithm and OMP for sparse coding. The computation is dominated by sparse coding which scales as $\mathcal{O}(nJLTB_s)$, where *T* is the number of iterations in learning and *J* is the size of sparsifying basis. The costs of sparse coding all *N* patches by adaptively learned sparsifying basis is $\mathcal{O}(nJLB_s)$. For a 256×256 image, the proposed algorithm requires about 8~9 minutes for CS recovery, on an Intel Core2 Duo 2.96G PC under Matlab R2011a environment. Finally, we provide the computational time comparisons with various algorithms for *Image* House at different CS *subrates* in Table 4.

*5.6 Algorithm Convergence*

Since the objective function (6) is non-convex, it is difficult to give its theoretical proof for global convergence. Here, we only provide empirical evidence to illustrate the good convergence of the proposed algorithm. Fig. 12 plots the evolutions of PSNR versus iteration numbers for four test images with various subrates (*subrate=30%* and *subrate=40%*). It is observed that with the growth of iteration number, all the PSNR curves increase monotonically and ultimately become flat and stable, exhibiting good convergence property. Note that due to non-convexity of Eq. (6), it is natural that there are some perturbations in the curves.

## 6. Conclusion

In this paper, we propose to characterize the intrinsic sparsity of natural images by patch-based redundant sparse representation using adaptively learned sparsifying basis. This particular type of spare representation is formulated by non-convex $\ell_0$ minimization for image compressive sensing recovery, which can be efficiently solved by the developed split Bregman iteration based technique. Experimental results on a wide range of natural images for CS recovery have shown that our proposed algorithm achieves significant performance improvements over many current state-of-the-art schemes and exhibits good convergence property.

**Table 1:** A Complete Description of Proposed Framework for Image CS Recovery

**Input:** the CS measurements $y$ and the random projection operator $\Phi$

**Initialization:** $t=0, u^{(0)}=0, b^{(0)}=0, \alpha^{(0)}=0, B_s, \eta, \omega, \lambda, \mu$;

**Repeat**

   Update $u^{(t+1)}$ by $u^{(t+1)} = u^{(t)} - \eta(\Phi^T\Phi u^{(t)} - \Phi^T y + \mu(u^{(t)} - D^{(t)} \circ \alpha^{(t)} - b^{(t)}))$;

   $r^{(t+1)} = u^{(t+1)} - b^{(t)}$; $\tau = (\lambda K)/(\mu N)$;

   Update $D^{(t+1)}$ by $D^{(t+1)} = \underset{D}{\operatorname{argmin}} \sum_{k=1}^{J} \left\| r_k^{(t+1)} - D\alpha_k \right\|_2^2$ s.t. $\left\|\alpha_k\right\|_p \leq L, \forall k$;

   $\delta = \omega\tau$;

   *for* Each patch $x_{G_k}$

       Reconstruct $\hat{\alpha}_k^{(t+1)}$ by computing $\hat{\alpha}_k^{(t+1)} = \operatorname{argmin}_{\alpha_k} \left\|\alpha_k\right\|_0$ s.t. $\left\|D\alpha_k - r_k\right\|_2^2 \leq \delta$;

   *end for*

   Update $\hat{\alpha}^{(t+1)}$ by concatenating all $\hat{\alpha}_k^{(t+1)}$;

   Update $b^{(t+1)}$ by computing $b^{(t+1)} = b^{(t)} - (u^{(t+1)} - D^{(t+1)} \circ \alpha^{(t+1)})$;

   $t \leftarrow t+1$;

**Until** maximum iteration number is reached

**Output:** Final restored image $\hat{x} = D \circ \hat{\alpha}$.

**Table 2:** PSNR Comparisons with Various CS Recovery Methods (Unit: dB)

| Subrate | Algorithms | House | Barbara | Leaves | Monarch | Parrot | Vessels | Avg. |
|---|---|---|---|---|---|---|---|---|
| | DWT | 30.70 | 23.96 | 22.05 | 24.69 | 25.64 | 21.14 | 24.70 |
| | TV | 31.44 | 23.79 | 22.66 | 26.96 | 26.6 | 22.04 | 25.59 |
| 20% | MH | 33.60 | 31.09 | 24.54 | 27.03 | 28.06 | 24.95 | 28.21 |
| | CoS | 34.34 | 26.60 | **27.38** | **28.65** | 28.44 | 26.71 | 28.69 |
| | **Proposed** | **35.86** | **31.61** | 27.15 | 28.23 | **29.56** | **30.14** | **30.43** |
| | DWT | 33.60 | 26.26 | 24.47 | 27.23 | 28.03 | 24.82 | 27.40 |
| | TV | 33.75 | 25.03 | 25.85 | 30.01 | 28.71 | 25.13 | 28.08 |
| 30% | MH | 35.54 | 33.47 | 27.65 | 29.18 | 31.20 | 29.36 | 31.07 |
| | CoS | 36.69 | 29.49 | 31.02 | 31.38 | 30.39 | 31.35 | 31.72 |
| | **Proposed** | **38.15** | **34.73** | **31.10** | **31.48** | **32.24** | **34.60** | **33.72** |
| | DWT | 35.69 | 28.53 | 26.82 | 29.58 | 30.06 | 29.53 | 30.03 |
| | TV | 35.56 | 26.56 | 28.79 | 32.92 | 30.54 | 28.14 | 30.42 |
| 40% | MH | 37.04 | 35.20 | 29.93 | 31.07 | 33.21 | 33.49 | 33.32 |
| | CoS | 38.46 | 32.76 | 33.87 | 33.98 | 32.55 | 33.95 | 34.26 |
| | **Proposed** | **40.13** | **37.16** | **34.66** | **34.33** | **34.38** | **38.27** | **36.49** |

**Table 3:** FSIM Comparisons with Various CS Recovery Methods

| Subrate | Algorithms | House | Barbara | Leaves | Monarch | Parrot | Vessels | Avg. |
|---|---|---|---|---|---|---|---|---|
| | DWT | 0.9029 | 0.8547 | 0.7840 | 0.8155 | 0.9161 | 0.8230 | 0.8494 |
| | TV | 0.9051 | 0.8199 | 0.8553 | 0.8870 | 0.9018 | 0.8356 | 0.8675 |
| 20% | MH | 0.9370 | 0.9419 | 0.8474 | 0.8707 | 0.9332 | 0.8756 | 0.9010 |
| | CoS | 0.9326 | 0.8742 | **0.9304** | **0.9171** | 0.9282 | 0.9214 | 0.9259 |
| | Proposed | **0.9542** | **0.9487** | 0.9106 | 0.8879 | **0.9433** | **0.9499** | **0.9324** |
| | DWT | 0.9391 | 0.8980 | 0.8314 | 0.8628 | 0.9445 | 0.8924 | 0.8947 |
| | TV | 0.9384 | 0.8689 | 0.9092 | 0.9279 | 0.9329 | 0.9011 | 0.9131 |
| 30% | MH | 0.9546 | 0.9614 | 0.8993 | 0.9003 | 0.9529 | 0.9360 | 0.9341 |
| | CoS | 0.9592 | 0.9267 | **0.9606** | **0.9449** | 0.9490 | 0.9664 | 0.9511 |
| | Proposed | **0.9722** | **0.9716** | 0.9509 | 0.9328 | **0.9622** | **0.9775** | **0.9612** |
| | DWT | 0.9576 | 0.9327 | 0.8741 | 0.9011 | 0.9588 | 0.9467 | 0.9285 |
| | TV | 0.9574 | 0.9088 | 0.9442 | 0.9538 | 0.9530 | 0.9441 | 0.9436 |
| 40% | MH | 0.9676 | 0.9727 | 0.9276 | 0.9217 | 0.9651 | 0.9677 | 0.9537 |
| | CoS | 0.9724 | 0.9618 | **0.9744** | **0.9637** | 0.9627 | 0.9784 | 0.9689 |
| | Proposed | **0.9817** | **0.9829** | 0.9736 | 0.9567 | **0.9735** | **0.9886** | **0.9762** |

**Table 4:** Computational Time Comparisons with Various Algorithms (Unit: s)

| Image | Subrate | DWT [30] | TV [29] | MH [42] | CoS [43] | Proposed |
|---|---|---|---|---|---|---|
| *House* (256×256) | *20%* | 12.6 | 9.9 | 21.6 | 315.9 | 354.5 |
| | *30%* | 8.1 | 8.1 | 46.7 | 245.6 | 339.1 |
| | *40%* | 5.9 | 7.5 | 27.2 | 216.8 | 330.4 |
| **Average** | | 8.9 | 8.5 | 31.9 | 259.5 | 341.7 |

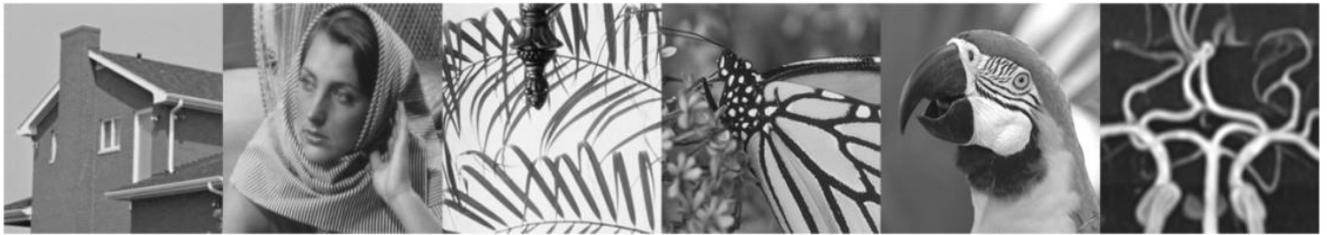

**Figure 1:** Six experimental test images.

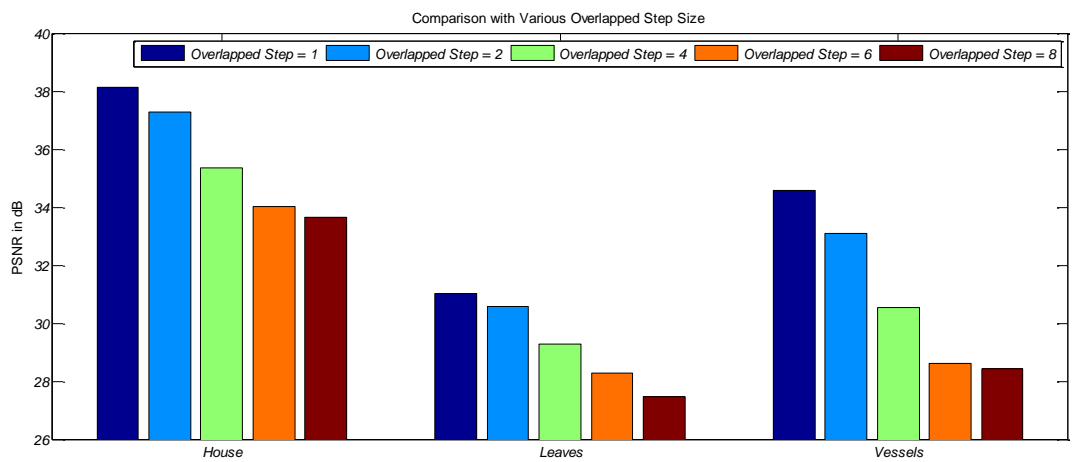

**Figure 2:** Comparison of image CS recovery with different overlapped step size for three test images in the case of *subrate=30%*.

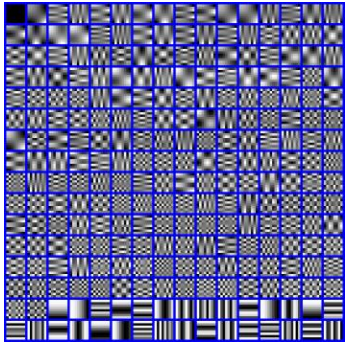 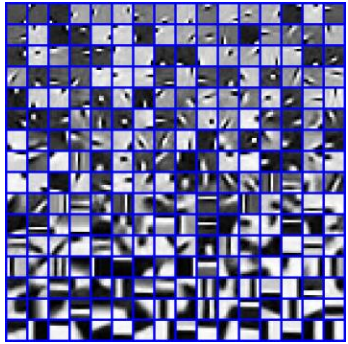 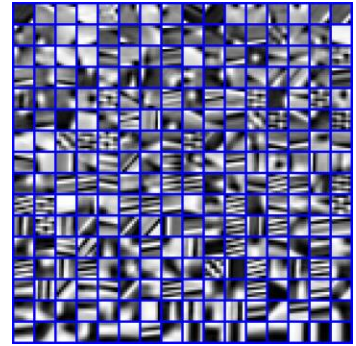

(a)              (b)              (c)

**Figure 3:** Display of three types of sparsifying basis. From left to right: the fixed over-complete DCT basis; the globally learned sparsifying basis from a large set of natural images; the adaptively learned sparsifying basis from the processed image at last iteration with respect to image House in the case of *subrate=20%*.

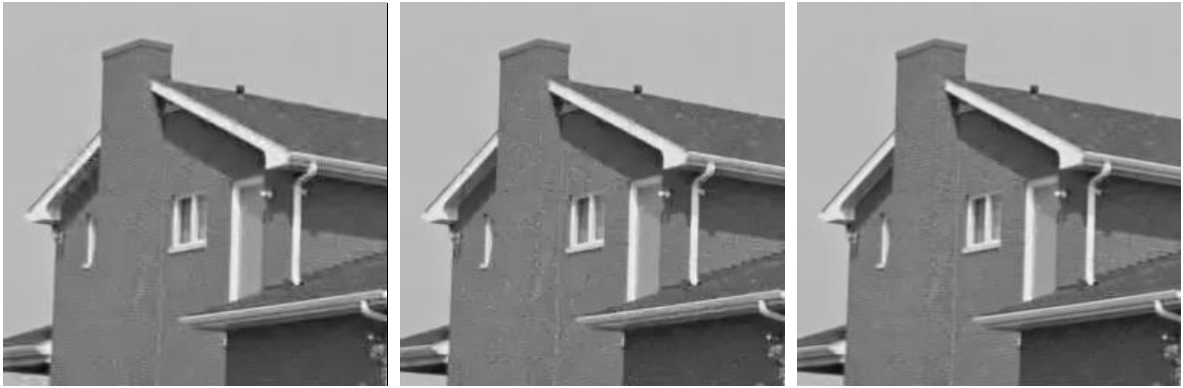

**Figure 4:** Visual comparison of image CS recovery using different types of sparsifying basis for image *House* in the case of *subrate=20%*. From left to right: the recovered image by over-complete DCT basis (PSNR=33.97dB); the recovered image by globally learned sparsifying basis (PSNR=34.62dB); the recovered image by the adaptively learned sparsifying basis at each iteration (PSNR=35.86dB).

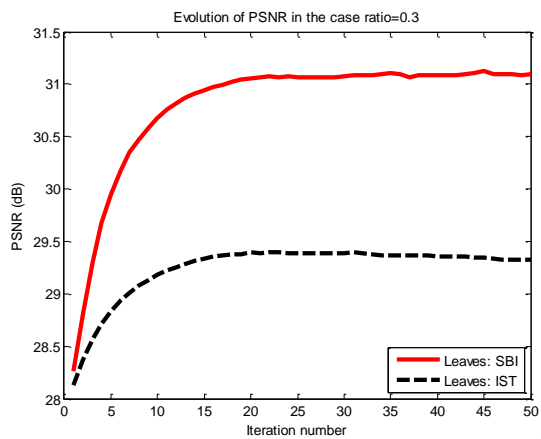 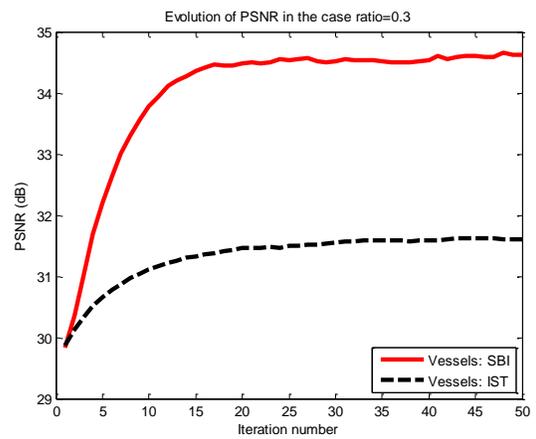

(a)                                         (b)

**Figure 5:** Comparison between SBI and IST for solving our proposed $\ell_0$ minimization Eq. (6). From left to right: progression of the PSNR (dB) results achieved by proposed $\ell_0$ minimization with respect to the iteration number for gray images *Leaves* and *Vessels* in the cases of image CS recovery with *subrate=30%*.

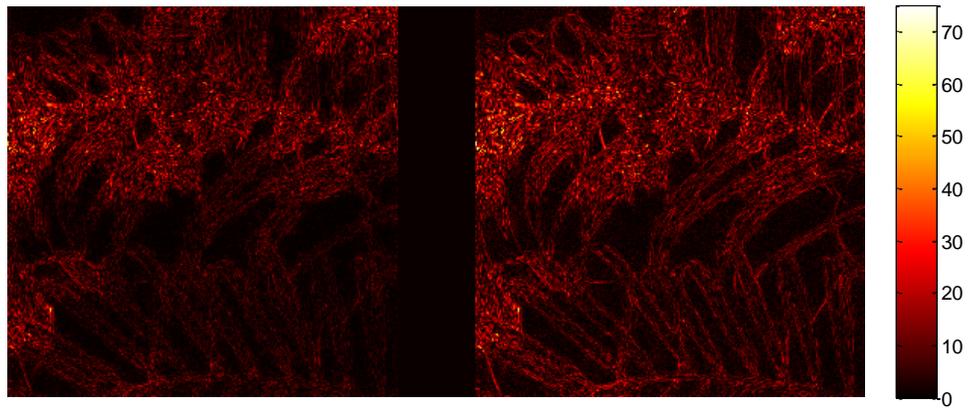

**Figure 6:** Visual comparison between SBI and IST for image CS recovery with respect to image *Leaves* in the case of *subrate=30%*. Left: magnitude of recovery error for SBI; Right: magnitude of recovery error for IST.

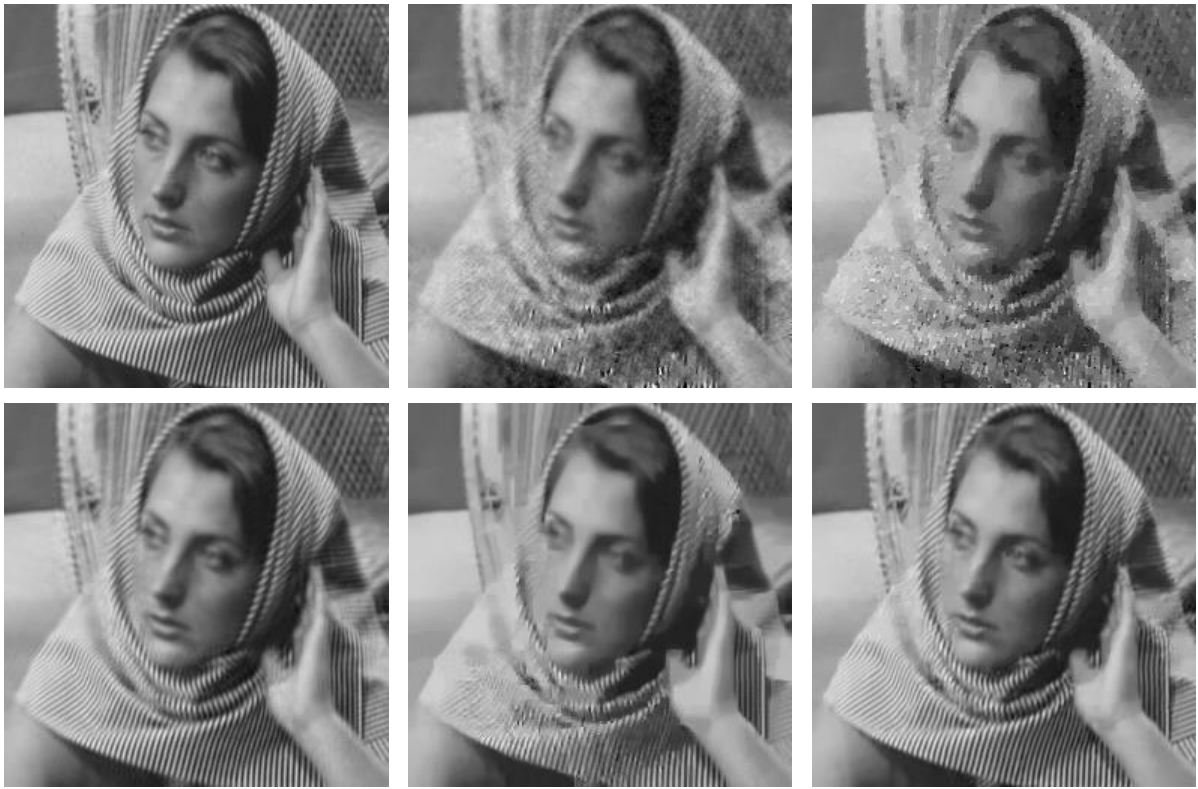

**Figure 7:** Visual quality comparison of image CS recovery on gray image Barbara in the case of *subrate=20%*. From left to right and top to bottom: original image, the CS recovered images by DWT (PSNR=23.96dB; FSIM=0.8547), TV (PSNR=23.79dB; FSIM =0.8199), MH (PSNR=31.09dB; FSIM=0.9419), CoS (PSNR=26.60dB; FSIM=0.8742) and the proposed algorithm (PSNR=31.61dB; FSIM =0.9487).

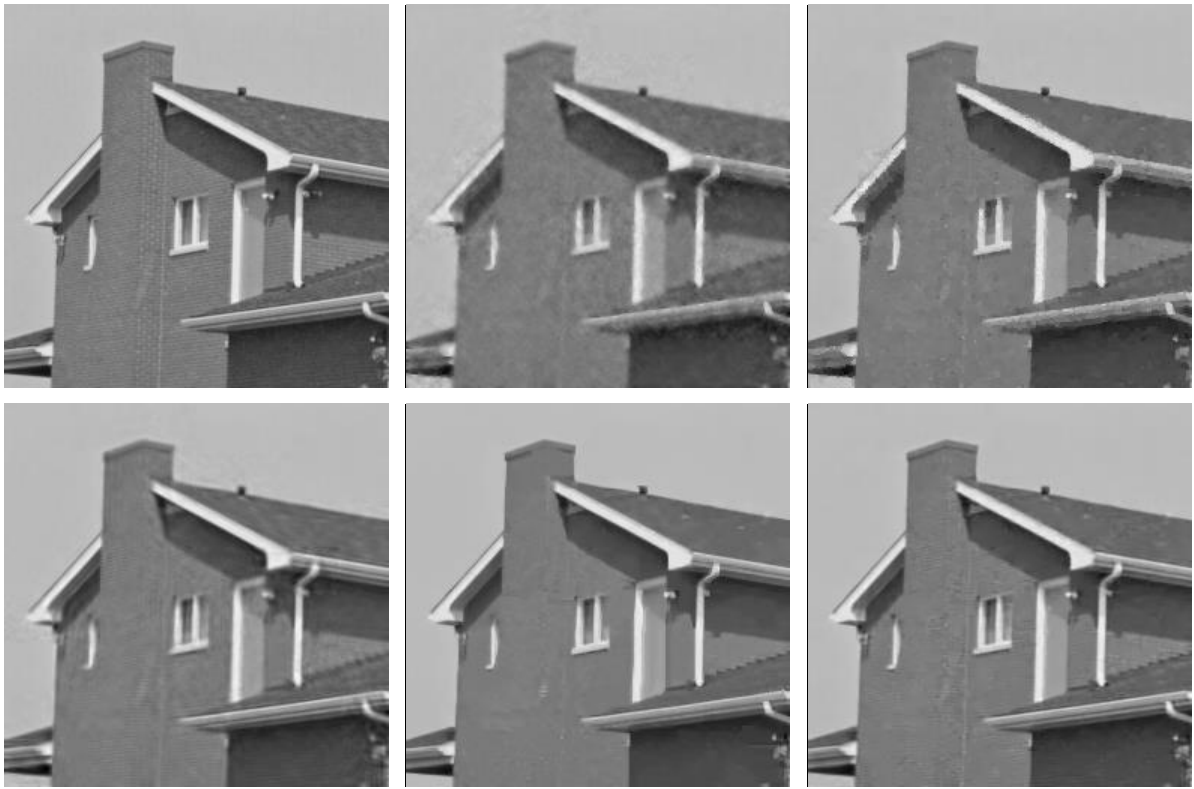

**Figure 8:** Visual quality comparison of image CS recovery on gray image *House* in the case of *subrate=20%*. From left to right and top to bottom: original image, the CS recovered images by DWT (PSNR=30.70dB; FSIM=0.9029), TV (PSNR=31.44dB; FSIM =0.9051), MH (PSNR=33.60dB; FSIM=0.9370), CoS (PSNR=34.34dB; FSIM=0.9326) and the proposed algorithm (PSNR=35.86dB; FSIM =0.9542).

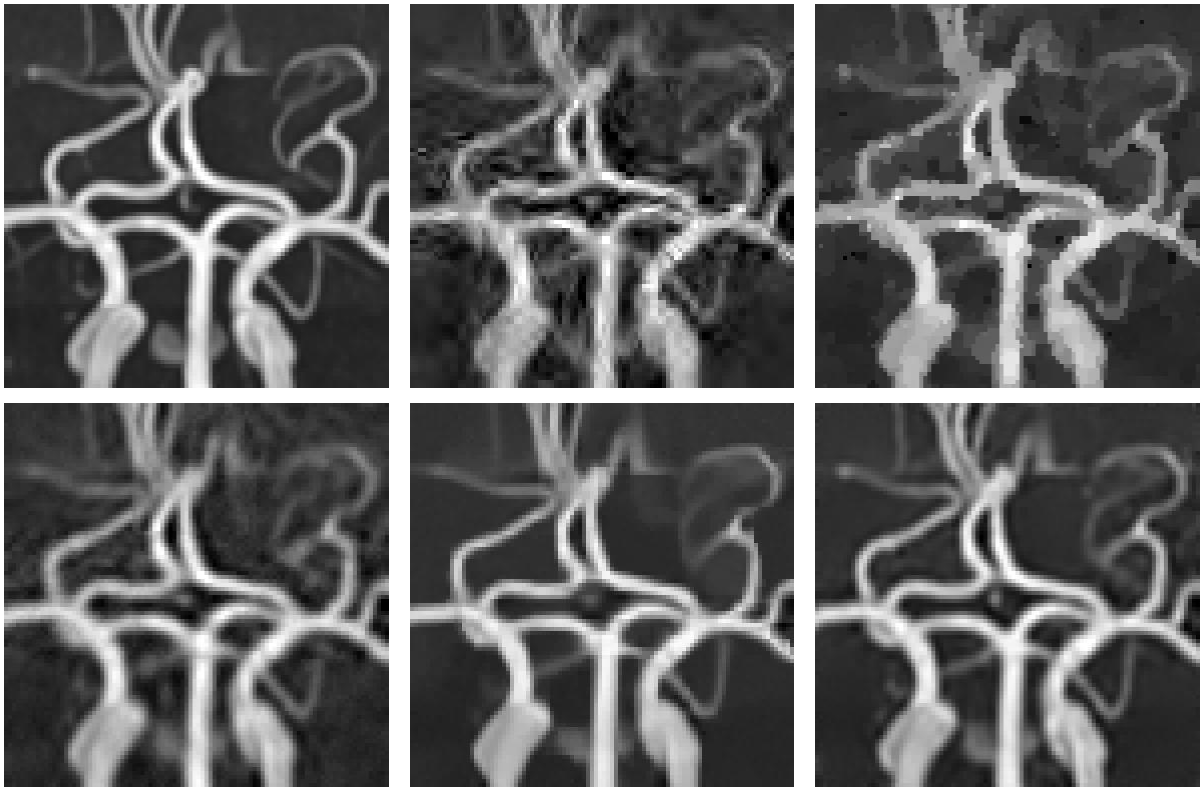

**Figure 9:** Visual quality comparison of image CS recovery on gray image *Vessels* in the case of *subrate=20%*. From left to right and top to bottom: original image, the CS recovered images by DWT (PSNR=21.14dB; FSIM=0.8230), TV (PSNR=22.04dB; FSIM =0.8356), MH (PSNR=24.95dB; FSIM=0.8756), CoS (PSNR=26.71dB; FSIM=0.9214) and the proposed algorithm (PSNR=30.14dB; FSIM =0.9499).

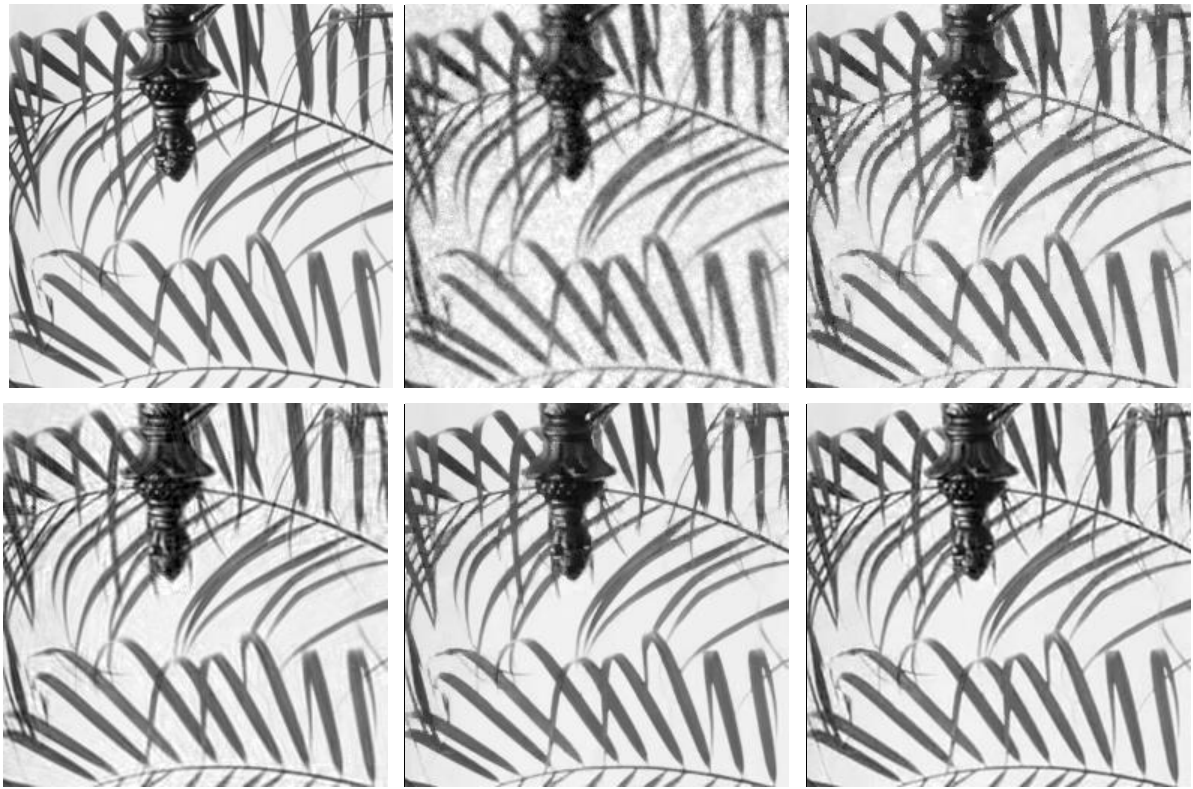

**Figure 10:** Visual quality comparison of image CS recovery on gray image *Leaves* in the case of *subrate=30%*. From left to right and top to bottom: original image, the CS recovered images by DWT (PSNR=24.47dB; FSIM=0.8314), TV (PSNR=25.85dB; FSIM =0.9092), MH (PSNR=27.65dB; FSIM=0.8993), CoS (PSNR=31.02dB; FSIM=0.9606) and the proposed algorithm (PSNR=31.10dB; FSIM =0.9509).

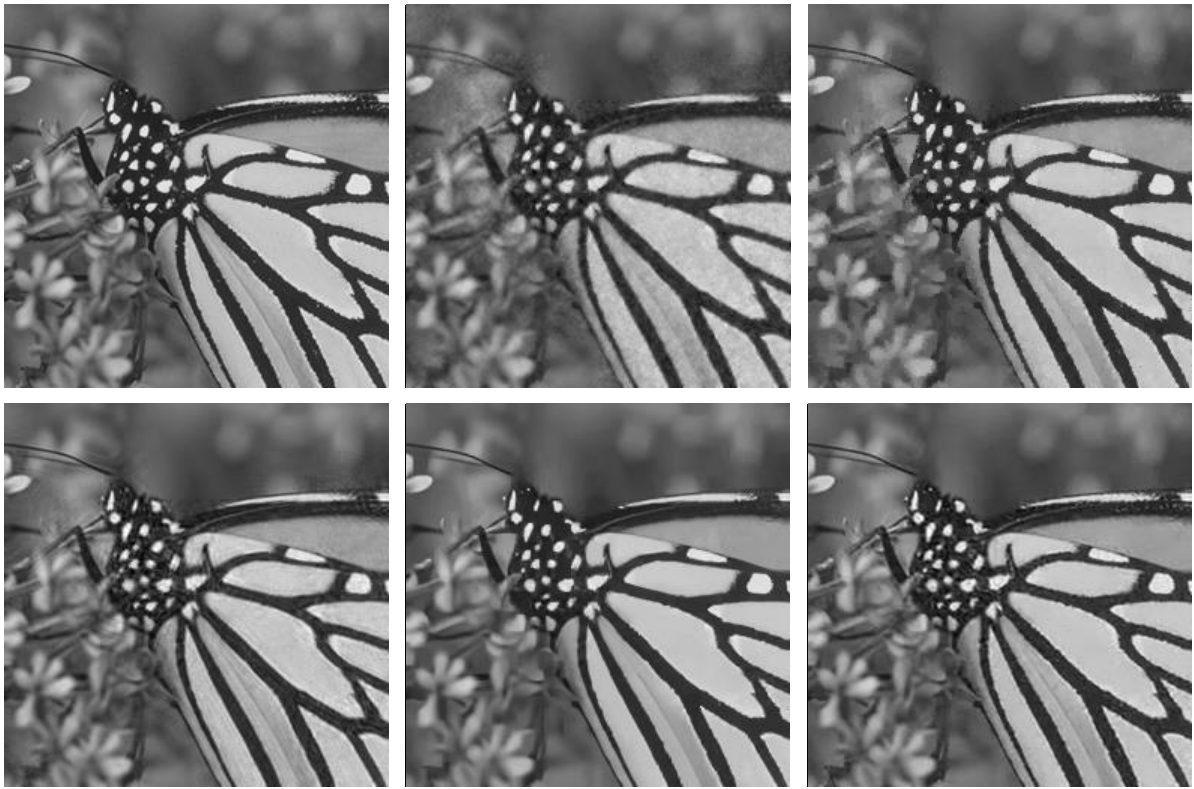

**Figure 11:** Visual quality comparison of image CS recovery on gray image *Monarch* in the case of *subrate=30%*. From left to right and top to bottom: original image, the CS recovered images by DWT (PSNR=27.23dB; FSIM=0.8628), TV (PSNR=30.01dB; FSIM =0.9279), MH (PSNR=29.18dB; FSIM=0.9003), CoS (PSNR=31.38dB; FSIM=0.9449) and the proposed algorithm (PSNR=31.48dB; FSIM =0.9328).

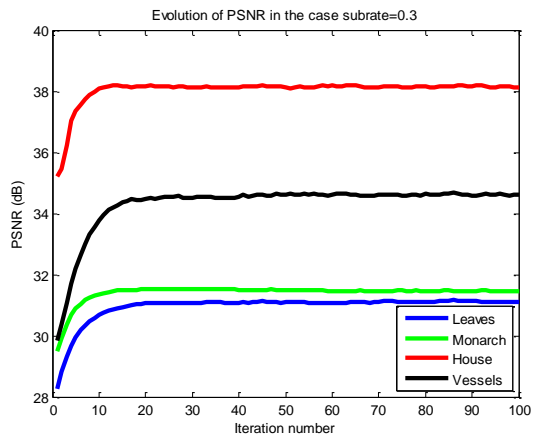 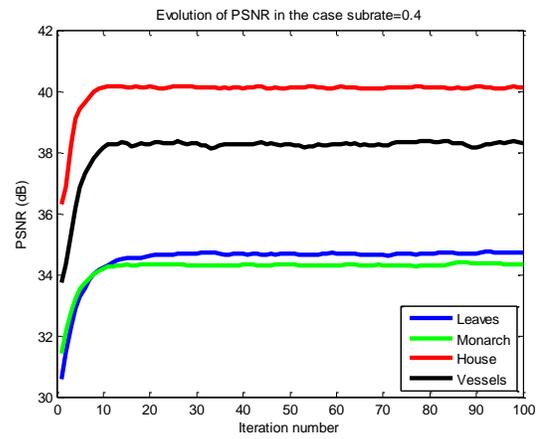

**Figure 12:** Convergence of the proposed algorithm. From left to right: Progression of the PSNR (dB) results achieved by proposed algorithm for four test images with respect to the iteration number in the cases of *subrate=30%* and *subrate=40%*.